%% file: main.tex
\documentclass[11pt]{article}

\usepackage[preprint]{acl}

\usepackage{times}
\usepackage{latexsym}

\usepackage[T1]{fontenc}

\usepackage[utf8]{inputenc}

\usepackage{microtype}

\usepackage{inconsolata}

\usepackage{graphicx}

\usepackage{amsmath}
\usepackage{amssymb}
\usepackage{comment}
\usepackage{xspace}
\usepackage{bm}
\usepackage{xcolor}
\usepackage{booktabs}
\usepackage{pifont}
\usepackage[most]{tcolorbox}
\usepackage[table]{xcolor}
\usepackage{array}
\usepackage{multirow}
\usepackage{amsmath}

\usepackage{algorithm}
\usepackage{algpseudocode}
\usepackage{etoc}        
\usepackage{changepage} 
\usepackage{hyperref}   
\usepackage{tabularx}
\usepackage{subcaption}

\newcommand{\ours}{\textsc{TraceSteer}\xspace}

\definecolor{findinggreen}{HTML}{005900}

\definecolor{baselineblue}{RGB}{249,246,243}
\definecolor{oursblue}{RGB}{219,237,255}
\definecolor{backbonebrown}{RGB}{244, 238, 228}

\definecolor{gainblue}{RGB}{45,180,190}
\definecolor{lossred}{RGB}{235,105,95}

\newcommand{\gain}[1]{%
    {\scriptsize\textcolor{gainblue}{\ensuremath{\uparrow}#1}}%
}

\newcommand{\loss}[1]{%
    {\scriptsize\textcolor{lossred}{\ensuremath{\downarrow}#1}}%
}

\newtcblisting{generationprompt}[1]{
    listing only,
    breakable,
    colback=blue!5,
    colframe=blue!50,
    title={#1},
    fonttitle=\bfseries,
    boxrule=0.5pt,
    arc=2mm,
    listing options={
        basicstyle=\ttfamily\footnotesize,
        breaklines=true,
        columns=fullflexible,
        keepspaces=true,
        showstringspaces=false
    }
}

\title{\ours: Trace Driven Activation Steering for Hallucination Mitigation in Diffusion Language Models}

\author{
  \textbf{Xin Zhang\textsuperscript{1,*}},
  \textbf{Yili Wang\textsuperscript{1,*}},
  \textbf{Yue Tan\textsuperscript{2}},
  \textbf{Xin He\textsuperscript{1}},
  \textbf{Rui Miao\textsuperscript{1}},\\
  \textbf{Yixin Liu\textsuperscript{2,\ensuremath{\dagger}}},
  \textbf{Yi Chang\textsuperscript{1}},
  \textbf{Xin Wang\textsuperscript{1,\ensuremath{\dagger}}}
\\
\\
  \textsuperscript{1}Jilin University, China \quad  
  \textsuperscript{2}Griffith University, Australia
}

\begin{document}

\maketitle
\begingroup
\renewcommand{\thefootnote}{\fnsymbol{footnote}}
\footnotetext[1]{Equal contribution.}
\footnotetext[2]{Corresponding authors.}
\endgroup

\begin{abstract}
Diffusion large language models (dLLMs) have emerged as a promising alternative to autoregressive LLMs, yet remain susceptible to hallucination. Activation steering offers a promising mitigation approach by guiding parallel token predictions through internal representations. However, existing steering methods largely overlook two characteristics of dLLMs: \textbf{dynamic denoising states and temporal hallucination risks}, which challenge predefined steering vectors and uniform intervention strengths. We propose \ours, a trace-driven activation steering framework that uses preceding denoising predictions as feedback to \textbf{jointly adapt steering direction and strength}. Specifically, \emph{Denoising Feedback Steering (DF-Steering)} constructs step-specific steering vectors from preceding predictions and current hidden states. \emph{Trajectory-Guided Steering Gate (TS-Gate)} regulates their strength using prediction uncertainty and the denoising timestep. With the backbone frozen, factual supervision and magnitude regularization jointly guide the learning of adaptive steering interventions, promoting factual reliability and stability throughout denoising. Experiments on two dLLMs show improved factual reliability on TruthfulQA and generalization across five downstream benchmarks. Our code is available at \url{https://github.com/zhangxincode/TRACESTEER}.
\end{abstract}

\etocdepthtag.toc{main}

\section{Introduction}

\input{paper/intro_v1}

\section{Preliminaries}
\input{paper/prel_v2}

\section{Methodology}
\input{paper/meth_v2}

\section{Experiment}

\input{paper/exp_v2}


\section{Conclusion}
In this work, we proposed \textbf{\ours}, a trace-driven activation steering framework for mitigating hallucinations in dLLMs. By leveraging intermediate generation traces, \ours dynamically adapts steering interventions throughout denoising to improve hallucination. Extensive experiments across multiple dLLM backbones and benchmarks demonstrate its broad utility and cross-task transferability. Our findings highlight the importance of incorporating denoising dynamics into activation steering for more reliable generation.
\clearpage

\section*{Limitations}

While \ours demonstrates effective hallucination mitigation for dLLMs, it still has certain limitations. In particular, our method relies on factual target responses to train the steering and gating modules. Such supervision provides explicit guidance for learning effective steering interventions, but also requires high-quality labeled data. However, obtaining reliable factual annotations can be expensive and time-consuming, especially in specialized domains. This dependence on supervised data may limit the applicability of our method to broader scenarios. In future work, we plan to explore weakly supervised and self-supervised learning strategies to reduce the reliance on factual annotations and improve the generalizability of our approach.

\section*{Ethical Considerations}

Our research involves no human subjects, animal experiments, or sensitive personal data. All experiments are conducted using publicly available models and benchmark datasets. The proposed method aims to improve the factual reliability of diffusion large language models by mitigating hallucinated content. While activation steering can modify model generation behavior, our work does not introduce new data collection or deployment-related risks beyond those generally associated with large language models. We are committed to following established research ethics and responsible AI practices throughout this study.

\bibliography{custom}

\appendix
\clearpage
\onecolumn

\etocdepthtag.toc{appendix}

\begingroup
\renewcommand{\contentsname}{Appendix Contents}
\hypersetup{hidelinks}

\large
\linespread{1.5}\selectfont

\etocsettagdepth{main}{none}
\etocsettagdepth{appendix}{subsection}

\begin{adjustwidth}{-0.5cm}{-0.5cm}
\tableofcontents
\end{adjustwidth}

\endgroup

\twocolumn


\input{paper/app_v2}

\end{document}

%% file: paper/intro_v1.tex
Diffusion large language models (dLLMs) have recently emerged as a promising alternative to autoregressive large language models (LLMs), demonstrating strong potential in a wide range of language understanding and generation tasks~\cite{nie2026large,bie2025llada2,ye2508dream}. 
Unlike the predominant autoregressive paradigm that generates text sequentially from left to right~\cite{liu2024deepseek,achiam2023gpt}, dLLMs generate text through iterative denoising. In dLLMs, generation starts from a masked sequence, with the model predicting masked tokens in parallel at each step and remasking selected tokens for subsequent prediction. Such a generation paradigm enables bidirectional context interaction and joint token refinement, facilitating global dependency modeling and efficient parallel generation~\cite{sahoo2024simple,zhong2026parallelism}. Despite their merits, dLLMs, like conventional LLMs, remain susceptible to the hallucination issue, posing the risk of generating incorrect or unsupported content that appears plausible. The hallucination issue undermines the reliability and trustworthiness of dLLMs, leading to potentially misleading outputs in real-world applications~\cite{qian2026dynhd,zhang2026demts}.


\begin{figure}[t]
\centering
\includegraphics[width=\linewidth]{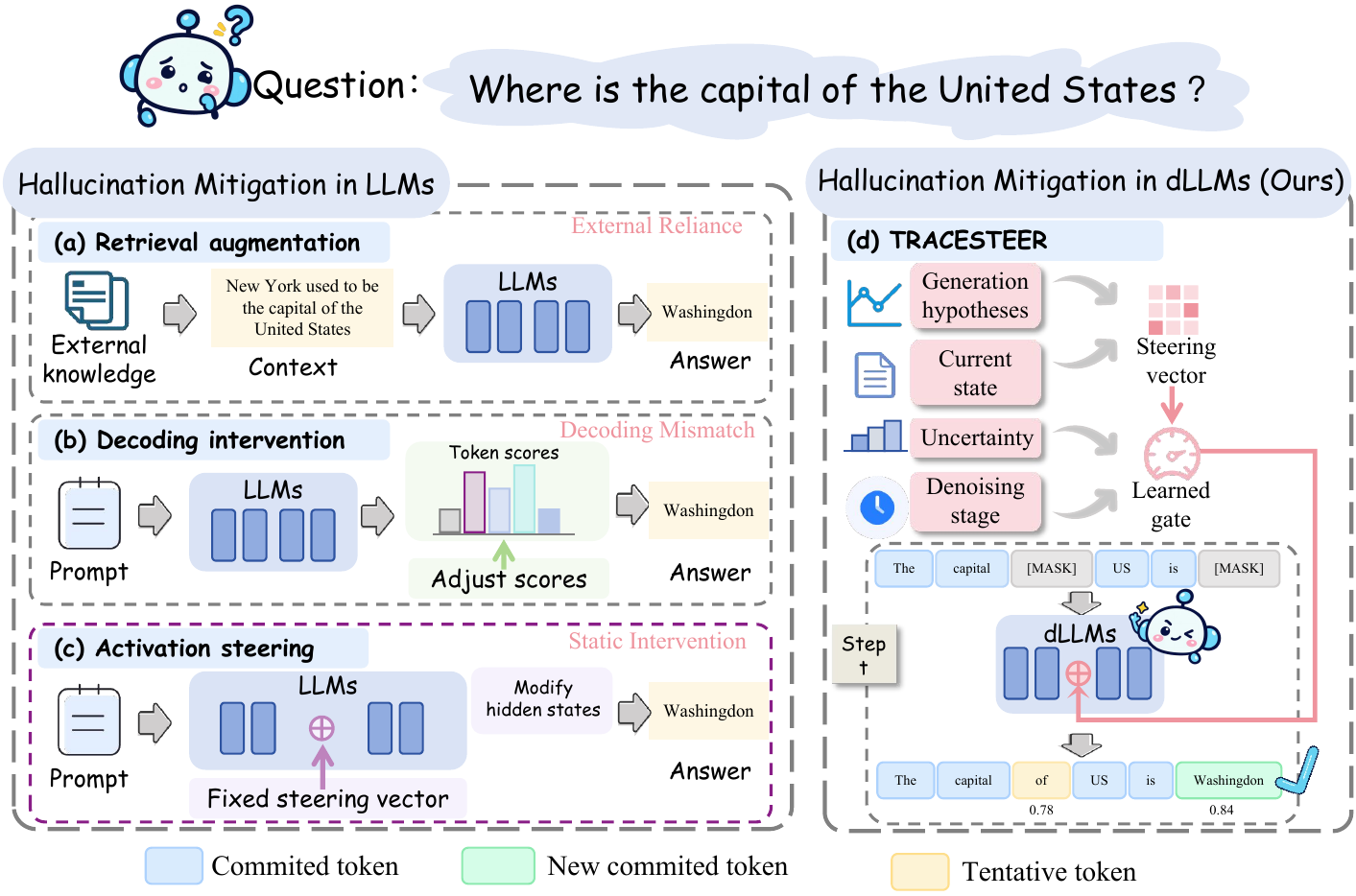}
\caption{\textbf{Motivation for trace-driven activation steering in dLLMs.} Existing methods in LLMs suffer from (a) reliance on external knowledge, (b) incompatibility with dLLMs' iterative denoising process, and (c) inability to adapt fixed steering vectors to evolving generation states. In contrast, (d) our approach enables dynamic, trace-aware activation steering.}
\vspace{-10pt}
\label{figure:introduction}
\end{figure}

To alleviate this issue, \textbf{hallucination mitigation} seeks to reduce hallucinated content and improve model reliability by intervening in generation. Nevertheless, most existing methods are developed for autoregressive LLMs and cannot be broadly applied to dLLMs~\cite{you2026spherical,su2025activation}. Specifically, \textit{retrieval augmentation} (Fig.~\ref{figure:introduction}a) leverages external knowledge to ground generation~\cite{gao2023retrieval,zhuang2026linearrag}; despite its effectiveness, maintaining external knowledge sources incurs additional costs and may not suit diverse scenarios.
\textit{Decoding intervention} (Fig.~\ref{figure:introduction}b) represents another line of methods that adjust token logits during generation~\cite{zhang2025alleviating,shi2024trusting}; however, they are designed for a single next-token decision and thus cannot directly accommodate dLLMs, where multiple token distributions are produced simultaneously and only a subset is selected for update. In contrast, \textit{\textbf{activation steering}} (Fig.~\ref{figure:introduction}c), which modulates model behavior by manipulating internal representations during inference, shows promising potential for hallucination mitigation in dLLMs~\cite{li2023inference,zhang2024truthx}. By directly manipulating internal representations, activation steering can guide multiple token predictions simultaneously, making it naturally compatible with diffusion-based parallel generation process.

Despite its promise, most activation steering techniques for hallucination mitigation are tailored to autoregressive LLMs, overlooking several unique characteristics of dLLM generation~\cite{rimsky2024steering,li2025hidden}. \textbf{Characteristic~\ding{182}: Dynamic denoising states.} dLLM generation evolves along a continuous denoising trajectory: at different denoising steps, the model not only exhibits varying generation states, but also continuously forms and revises its generation hypotheses.~\cite{wang2026tacg,zhong2026beyond} For example, when answering ``\textit{Why did the Roman Empire fall}'', some denoising steps may need to emphasize military pressure, while others shift toward political fragmentation. Facing such dynamic generation states, existing steering methods typically reuse a predefined steering vector throughout inference, which may fail to provide context-adaptive guidance along the evolving denoising trajectory. \textbf{Characteristic~\ding{183}: Temporal hallucination risks.} Different stages of dLLM generation play distinct roles and exhibit different sources of hallucination. Early denoising steps establish coarse generation hypotheses and are more prone to semantic errors, while later steps refine generated content, where excessive intervention may disrupt reliable predictions.~\cite{chang2026tracedet,weng2026tre} In this case, applying a uniform intervention strength throughout generation, as in existing steering methods, may fail to meet the varying steering needs at different generation stages, leading to over-steering or under-steering for hallucination mitigation.



Targeting the above characteristics, we propose \textbf{\ours} (Fig.~\ref{figure:introduction}d), a trace-driven activation steering framework for hallucination mitigation in dLLMs. \ours uses the trajectory information as feedback to jointly adapt the steering direction and intervention strength. To capture \textbf{Characteristic~\ding{182}}, we introduce a \emph{Denoising Feedback Steering (DF-Steering)} mechanism, which learns adaptive steering signals for different denoising states. Specifically, \ours encodes the preceding tentative response and relates its representations to the current hidden states. Trained under factual supervision, this mechanism learns a step-specific steering vector conditioned on both the generation hypotheses and the current generation state.
To model \textbf{Characteristic~\ding{183}}, we further introduce a \emph{Trajectory-Guided Steering Gate (TS-Gate)} mechanism that adjusts the intervention strength according to prediction uncertainty and generation stage. To be concrete, \ours learns a gating function conditioned on the uncertainty of the preceding prediction and the current denoising timestep, allowing the steering strength to adapt throughout the generation trajectory. 
The two mechanisms are trained jointly with factual target responses and magnitude regularization while keeping the dLLM backbone frozen. Experiments on LLaDA-8B-Instruct and Dream-7B-Instruct demonstrate improved factual reliability on TruthfulQA and generalization across five downstream benchmarks.


%% file: paper/prel_v2.tex
\begin{figure*}[t]
    \centering
    \includegraphics[width=\textwidth]{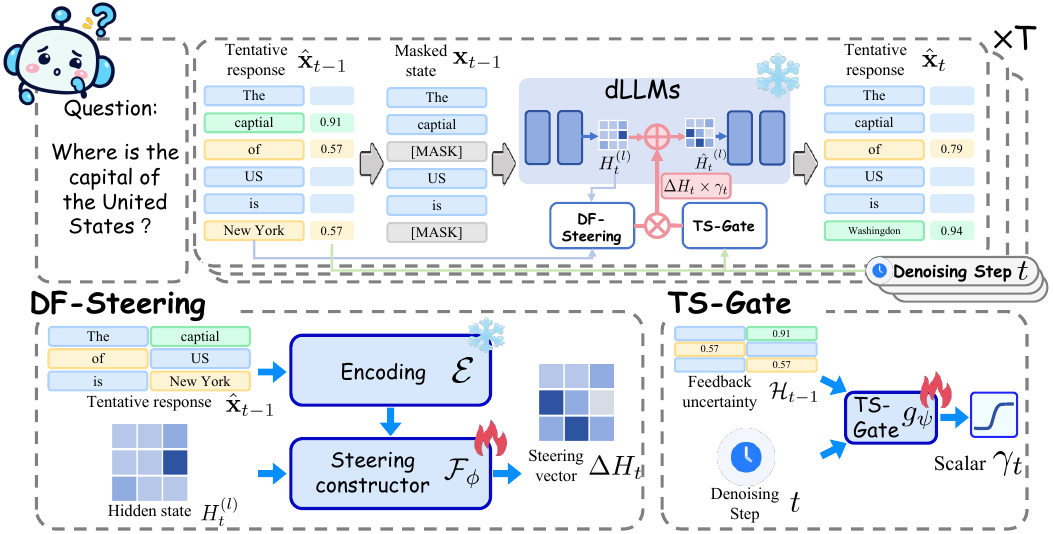}
    \caption{Overview of \ours. DF-Steering generates a feedback-guided steering vector from generation hypotheses and tentative predictions, while TS-Gate controls its strength based on the denoising timestep and feedback uncertainty. The gated steering vector is injected into dLLMs to mitigate hallucinations during denoising.}
    \label{fig:framework}
    \vspace{-10pt}
\end{figure*}

\subsection{Diffusion Language Model Generation}
\label{sec:dllm_generation}

We consider masked dLLMs with confidence-based decoding. Given an input prompt $c$, generation starts from a fully masked response $\mathbf{x}_0$ and proceeds through $T$ denoising iterations to produce the final response $\mathbf{x}_T$. Here, $t\in\{0,\ldots,T-1\}$ indexes generation iterations, and $\mathbf{x}_t=[x_t^1,\ldots,x_t^L]$ denotes the input state at iteration $t$, with $x_t^i\in\mathcal{V}\cup\{[\mathrm{MASK}]\}$, where $L$ is the response length and $\mathcal{V}$ is the vocabulary.

At iteration $t$, the backbone $\Phi$ predicts the masked positions $\mathcal{M}_t=\{i\mid x_t^i=[\mathrm{MASK}]\}$ in parallel. Let $p_t^i(v)=p_{\Phi}(x^i=v\mid c,\mathbf{x}_{t})$ denote the token probability at position $i$. Combining these predictions with the committed tokens yields a complete tentative response $\hat{\mathbf{x}}_t$, defined as
\begin{equation}
\hat{x}_{t+1}^i =
\begin{cases}
\arg\max_{v\in\mathcal{V}}p_t^i(v), & i\in\mathcal{M}_t,\\
x_t^i, & i\notin\mathcal{M}_t.
\end{cases}
\end{equation}

The model then commits a subset of masked positions with the highest prediction confidence according to the denoising schedule. The remaining tentative predictions are remasked, while previously committed tokens are preserved, producing $\mathbf{x}_{t+1}$. This process repeats until all positions are committed. Thus, each iteration produces both a complete tentative response $\hat{\mathbf{x}}_{t+1}$ and a potentially partially masked state $\mathbf{x}_{t+1}$. The tentative responses across iterations form a generation trace, capturing the model's evolving predictions before all tokens are committed.

\subsection{Hallucination Mitigation in dLLMs}
\label{sec:hallucination_mitigation}

Hallucination mitigation in dLLMs aims to reduce factually incorrect content in the final response $\mathbf{x}_T$ while preserving its relevance and informativeness. We study this task through interventions during denoising, seeking to improve the factuality of token predictions before they are committed to the response.

We consider a supervised setting with a training set $\mathcal{D}=\{(c^{(n)},\mathbf{y}^{(n)})\}_{n=1}^{N}$, where $\mathbf{y}^{(n)}$ denotes the factual target response for prompt $c^{(n)}$. These targets provide supervision for learning the intervention. At inference time, the intervention has access to the prompt $c$, the current denoising state $\mathbf{x}_t$, and tentative predictions from preceding iterations, but not the factual target $\mathbf{y}$. It must therefore guide factual correction using the model's own generation information. A summary of related works is given in Appendix \ref{appendix:relatedwork}.

%% file: paper/meth_v2.tex
Building on the generation trace defined in Section~\ref{sec:dllm_generation}, we present \ours, a trace-driven activation steering framework that uses preceding denoising predictions to guide the current generation. At each denoising iteration, the backbone computes hidden representations of the response tokens and transforms them through successive Transformer layers to produce token predictions. As illustrated in Fig.~\ref{fig:framework}, \ours constructs an intervention from both the preceding denoising prediction and the current hidden states, and applies it at an intermediate layer. Its strength is regulated according to prediction uncertainty and denoising progress. Let $H_t^{(l)}\in\mathbb{R}^{L\times d}$ denote the response hidden states at layer $l$ and iteration $t$, where $d$ is the hidden dimension. We intervene as follows:
\begin{equation}
\widetilde{H}_t^{(l)}
=
H_t^{(l)}
+
\gamma_t\Delta H_t,
\label{eq:steering_intervention}
\end{equation}
where, for $t\geq1$, $\Delta H_t\in\mathbb{R}^{L\times d}$ is the steering vector constructed by \textit{DF-Steering} from prediction feedback and the current hidden states, and $\gamma_t\in(0,1)$ is the intervention strength determined by TS-Gate. At $t=0$, no preceding prediction is available, so we leave the hidden states unchanged, i.e., $\widetilde{H}_0^{(l)}=H_0^{(l)}$. The resulting token predictions provide feedback for the next iteration. For subsequent iterations, the modified hidden states are propagated through the remaining backbone layers to produce the current token predictions. Sections~\ref{sec:dynamic_steering} and~\ref{sec:steering_gating} detail DF-Steering and TS-Gate, respectively. Both mechanisms are trained jointly under factual supervision while keeping the dLLM backbone frozen.

\subsection{Denoising Feedback Steering}
\label{sec:dynamic_steering}

Unlike a predefined steering vector reused throughout generation, steering in dLLMs needs to account for the generation hypotheses formed and revised along the denoising trajectory. The preceding tentative response $\hat{\mathbf{x}}_{t-1}$ reveals the model's latest generation hypothesis, but how this feedback should guide the current prediction depends on the current denoising state $H_t^{(l)}$. We therefore introduce \emph{Denoising Feedback Steering (DF-Steering)} to construct a step-specific steering vector by relating the preceding prediction to the current hidden states. \textit{DF-Steering} first encodes the preceding tentative response into contextual representations, then integrates them with the current hidden states to generate a steering vector under factual supervision.

\paragraph{Prediction Feedback Encoding.}
The preceding tentative response exposes the model's predicted content, including predictions at positions that remain masked in the current input $\mathbf{x}_t$. To use this content for activation steering, we encode it into contextual representations at the same layer as the hidden states to be modified. For $t\geq1$, we concatenate the prompt $c$ with the complete tentative response $\hat{\mathbf{x}}_{t-1}$ and pass the sequence through the frozen backbone up to the intervention layer $l$. Let $\mathcal{E}$ denote this encoding operation, retaining only the hidden states at response positions:
\begin{equation}
Z_{t-1}
=
\mathcal{E}\left(c,\hat{\mathbf{x}}_{t-1}\right),
\label{eq:feedback_encoding}
\end{equation}
where $Z_{t-1}\in\mathbb{R}^{L\times d}$ contains the layer-$l$ representations of the preceding tentative response in the context of the prompt. This encoding makes the preceding prediction available as contextual feedback for constructing the intervention.

\paragraph{Steering Vector Construction.}
\textit{DF-Steering} translates the preceding generation hypothesis into a corrective direction for the current denoising state. Rather than directly injecting the encoded prediction, it learns how the current hidden states should draw on this feedback to shape the intervention. Specifically, each response position establishes state-dependent associations with the preceding prediction representations and aggregates their information into a position-specific steering signal. We formulate this feedback-to-steering transformation as:
\begin{equation}
\Delta H_t
=
\mathcal{F}_{\phi}
\left(H_t^{(l)},Z_{t-1}\right),
\label{eq:dynamic_steering}
\end{equation}
where $\mathcal{F}_{\phi}$ is a learnable steering constructor parameterized by $\phi$, e.g., a cross-attention mechanism. The current hidden states determine how prediction feedback is incorporated, while factual supervision shapes the resulting activation updates through their effects on subsequent token predictions. This formulation makes the steering direction responsive to both the model's preceding hypothesis and its current denoising state. Recomputing $\Delta H_t$ at each iteration $t\geq1$ allows the intervention to evolve with the generation trajectory.

\subsection{Trajectory-Guided Steering Gate}
\label{sec:steering_gating}

While \textit{DF-Steering} adapts the intervention direction to the current generation, denoising stages impose different demands on intervention strength. Early iterations form coarse generation hypotheses, where insufficient steering may leave semantic errors uncorrected. Later iterations refine increasingly committed content, where excessive steering may disturb reliable predictions. Applying a uniform strength across these stages risks both under-steering and over-steering. We introduce \textit{Trajectory-Guided Steering Gate (TS-Gate)} to learn how strongly to intervene from the denoising stage and the uncertainty of the preceding prediction.

\paragraph{Feedback Uncertainty.}
The denoising timestep identifies the generation stage, but does not capture how uncertain the model's predictions are within a particular trajectory. Predictions at the same stage may exhibit different levels of uncertainty, calling for intervention strengths that account for the observed prediction state as well as generation progress. For $t\geq1$, we characterize this state using the predictive entropy from the preceding iteration, averaged over its masked positions:
\begin{equation}
\mathcal{H}_{t-1}
=
-\frac{1}{|\mathcal{M}_{t-1}|}
\sum_{i\in\mathcal{M}_{t-1}}
\sum_{v\in\mathcal{V}}
p_{t-1}^{i}(v)\log p_{t-1}^{i}(v),
\label{eq:prediction_entropy}
\end{equation}
where $\mathcal{M}_{t-1}$ denotes the masked positions in $\mathbf{x}_{t-1}$. This quantity summarizes uncertainty in the predictions used to form $\hat{\mathbf{x}}_{t-1}$, providing a trajectory-specific signal beyond the timestep alone.

\paragraph{Intervention Strength Regulation.}
TS-Gate relates this uncertainty signal to the current denoising stage, allowing similar uncertainty to produce different intervention strengths depending on when correction is applied. We learn this relationship through:
\begin{equation}
\gamma_t
=
\sigma\left(
g_{\psi}\left(t,\mathcal{H}_{t-1}\right)
\right),
\label{eq:steering_gate}
\end{equation}
where $g_{\psi}$ is a learnable gating function parameterized by $\psi$, implemented as an MLP, and $\sigma$ is the sigmoid function. The scalar $\gamma_t$ scales $\Delta H_t$ in Eq.~\ref{eq:steering_intervention}. Under factual supervision, the gate learns to balance corrective intervention with preservation of reliable predictions across denoising stages, without imposing a fixed strength schedule. \textit{DF-Steering} determines the state-specific steering direction, while TS-Gate regulates its influence along the generation trajectory.

\subsection{Learning to Steer with Factual Supervision}
\label{sec:training_objective}

To guide the model toward more factual generation, we jointly train DF-Steering and TS-Gate using factual target responses, learning both how to construct steering vectors and how strongly to apply them at each denoising stage. We further introduce magnitude regularization to constrain the steering vectors relative to the backbone representations, promoting more stable interventions.

\paragraph{Factual Supervision.}
Given a training pair $(c,\mathbf{y})\in\mathcal{D}$, where $\mathbf{y}=[y^1,\ldots,y^L]$ is the factual target response, we obtain the current denoising state $\mathbf{x}_t$ and the preceding tentative response $\hat{\mathbf{x}}_{t-1}$ from the denoising process at iteration $t\geq1$. Applying Eq.~\ref{eq:steering_intervention} and propagating the modified hidden states through the remaining backbone layers yields token probabilities $\widetilde{p}_t^i(v)$. We supervise these predictions at the currently masked positions:
\begin{equation}
\mathcal{L}_{\mathrm{task}}
=
-\frac{1}{|\mathcal{M}_t|}
\sum_{i\in\mathcal{M}_t}
\log \widetilde{p}_t^i(y^i),
\label{eq:task_loss}
\end{equation}
where $\mathcal{M}_t$ is the set of masked positions in $\mathbf{x}_t$. Supervising these positions directs learning toward predictions that remain eligible for commitment. Gradients through the modified hidden states jointly optimize DF-Steering's steering constructor and TS-Gate's regulation of intervention strength. The preceding prediction supplies the generation context for the intervention, while the factual target determines the correction objective.

\paragraph{Magnitude Regularization.}
Factual supervision encourages corrective updates, but does not explicitly constrain their magnitude. We therefore penalize the steering vector relative to the current hidden states, discouraging disproportionately large updates while accounting for variation in representation scale:
\begin{equation}
\mathcal{L}_{\mathrm{reg}}
=
\frac{\|\Delta H_t\|_F^2}
{\|H_t^{(l)}\|_F^2+\epsilon},
\label{eq:steering_regularization}
\end{equation}
where $\|\cdot\|_F$ denotes the Frobenius norm and $\epsilon$ is a small constant for numerical stability. The overall training loss combines factual supervision with magnitude regularization:
\begin{equation}
\mathcal{L}_{\mathrm{train}}
=
\mathcal{L}_{\mathrm{task}}
+
\lambda\mathcal{L}_{\mathrm{reg}},
\label{eq:overall_objective}
\end{equation}
where $\lambda$ controls the regularization strength. The trainable parameters are $\Theta=\{\phi,\psi\}$, corresponding to the feedback-to-steering mapping in \textit{DF-Steering} and the gating function in \textit{TS-Gate}, respectively. All backbone parameters remain frozen, while gradients propagate through the remaining backbone layers to optimize $\Theta$. At inference time, \ours constructs and regulates interventions using denoising feedback without access to factual target responses. Algorithm~\ref{alg:tracesteer} summarizes the procedure of \ours in Appendix~\ref{app:inference_algorithm}

%% file: paper/exp_v2.tex
\begin{table*}[t]
    \centering
    \caption{Performance comparison on TruthfulQA.
    Colored values indicate changes relative to the vanilla baseline.
    Best results are shown in \textbf{bold}, and second-best results are \underline{underlined}.}
    \label{tab:truthfulqa}

    \setlength{\tabcolsep}{6pt}
    \renewcommand{\arraystretch}{1.15}

    \resizebox{\linewidth}{!}{
    \begin{tabular}{clllllll}
        \toprule

        \multirow[c]{2}{*}{\textbf{Method}}
        & \multicolumn{4}{c}{\textbf{Multiple-Choice}}
        & \multicolumn{3}{c}{\textbf{Open-ended Generation}} \\

        \cmidrule(lr){2-5}
        \cmidrule(lr){6-8}

        & \textbf{MC1} ($\uparrow$)
        & \textbf{MC2} ($\uparrow$)
        & \textbf{MC3} ($\uparrow$)
        & \textbf{Avg.} ($\uparrow$)
        & \textbf{TRUE} ($\uparrow$)
        & \textbf{INFO} ($\uparrow$)
        & \textbf{TRUE$\times$INFO} ($\uparrow$) \\

        \midrule


        \rowcolor{backbonebrown}
        \multicolumn{8}{l}{\textbf{LLaDA-8B-Instruct}} \\

        \cellcolor{baselineblue}Vanilla
        & \cellcolor{baselineblue}22.13
        & \cellcolor{baselineblue}47.08
        & \cellcolor{baselineblue}23.45
        & \cellcolor{baselineblue}30.89
        & \cellcolor{baselineblue}49.18
        & \cellcolor{baselineblue}81.96
        & \cellcolor{baselineblue}40.31 \\

        Few-shot ICL
        & \underline{30.32}\gain{8.19}
        & 46.35\loss{0.73}
        & \underline{24.60}\gain{1.15}
        & \underline{33.76}\gain{2.87}
        & 45.08\loss{4.10}
        & 79.50\loss{2.46}
        & 35.84\loss{4.47} \\

        Oscar
        & 23.77\gain{1.64}
        & 46.97\loss{0.11}
        & 23.66\gain{0.21}
        & 31.47\gain{0.58}
        & 51.18\gain{2.00}
        & 84.32\gain{2.36}
        & 43.15\gain{2.84} \\

        CAA
        & 23.77\gain{1.64}
        & \underline{49.79}\gain{2.71}
        & 24.52\gain{1.07}
        & 32.69\gain{1.80}
        & 50.00\gain{0.82}
        & \underline{84.42}\gain{2.46}
        & 42.21\gain{1.90} \\

        L2S
        & 22.95\gain{0.82}
        & 46.96\loss{0.12}
        & 23.83\gain{0.38}
        & 31.25\gain{0.36}
        & \underline{52.45}\gain{3.27}
        & 82.78\gain{0.82}
        & \underline{43.42}\gain{3.11} \\

        \cellcolor{oursblue}\textbf{Ours}
        & \cellcolor{oursblue}\textbf{43.44}\gain{21.31}
        & \cellcolor{oursblue}\textbf{61.19}\gain{14.11}
        & \cellcolor{oursblue}\textbf{31.63}\gain{8.18}
        & \cellcolor{oursblue}\textbf{45.42}\gain{14.53}
        & \cellcolor{oursblue}\textbf{57.37}\gain{8.19}
        & \cellcolor{oursblue}\textbf{89.34}\gain{7.38}
        & \cellcolor{oursblue}\textbf{51.25}\gain{10.94} \\

        \midrule


        \rowcolor{backbonebrown}
        \multicolumn{8}{l}{\textbf{Dream-7B-Instruct}} \\

        \cellcolor{baselineblue}Vanilla
        & \cellcolor{baselineblue}24.59
        & \cellcolor{baselineblue}49.92
        & \cellcolor{baselineblue}24.31
        & \cellcolor{baselineblue}32.94
        & \cellcolor{baselineblue}49.18
        & \cellcolor{baselineblue}70.49
        & \cellcolor{baselineblue}34.67 \\

        Few-shot ICL
        & \underline{31.14}\gain{6.55}
        & 48.77\loss{1.15}
        & 25.67\gain{1.36}
        & 35.19\gain{2.25}
        & 48.47\loss{0.71}
        & 59.83\loss{10.66}
        & 29.00\loss{5.67} \\

        Oscar
        & 26.22\gain{1.63}
        & 51.24\gain{1.32}
        & 26.71\gain{2.40}
        & 34.72\gain{1.78}
        & 51.82\gain{2.64}
        & 72.59\gain{2.10}
        & 37.62\gain{2.95} \\

        CAA
        & 25.40\gain{0.81}
        & 50.41\gain{0.49}
        & 24.59\gain{0.28}
        & 33.47\gain{0.53}
        & \underline{54.09}\gain{4.91}
        & \underline{74.59}\gain{4.10}
        & \underline{40.35}\gain{5.68} \\

        L2S
        & 28.68\gain{4.09}
        & \underline{54.10}\gain{4.18}
        & \textbf{29.82}\gain{5.51}
        & \underline{37.53}\gain{4.59}
        & 53.82\gain{4.64}
        & 71.05\gain{0.56}
        & 38.24\gain{3.57} \\

        \cellcolor{oursblue}\textbf{Ours}
        & \cellcolor{oursblue}\textbf{36.88}\gain{12.29}
        & \cellcolor{oursblue}\textbf{55.95}\gain{6.03}
        & \cellcolor{oursblue}\underline{29.54}\gain{5.23}
        & \cellcolor{oursblue}\textbf{40.79}\gain{7.85}
        & \cellcolor{oursblue}\textbf{58.19}\gain{9.01}
        & \cellcolor{oursblue}\textbf{83.60}\gain{13.11}
        & \cellcolor{oursblue}\textbf{48.65}\gain{13.98} \\

        \bottomrule
    \end{tabular}
    }
    \vspace{-1pt}
\end{table*}

\label{sec:experiments}

Our experiments address three research questions:
\textbf{RQ1:} How effectively does \ours mitigate hallucinations in dLLMs while preserving response informativeness?
\textbf{RQ2:} How do prediction feedback, trajectory-guided gating, and magnitude regularization contribute to its effectiveness?
\textbf{RQ3:} Can \ours generalize across diverse tasks with a single set of learned parameters?

\subsection{Experimental Setup}
\label{sec:experimental_setup}

\paragraph{Backbones and Baselines.}
We evaluate \ours on LLaDA-8B-Instruct~\cite{nie2026large} and Dream-7B-Instruct~\cite{ye2508dream}. We compare against representative prompting-based methods (Few-shot ICL~\cite{liu2022few}), decoding-time methods (OSCAR~\cite{shah2026oscar}), and activation-steering methods (CAA~\cite{rimsky2024steering} and L2S~\cite{parekh2026learning}). We also report each backbone's original performance without intervention as Vanilla. More experimental setups can be found in Appendix ~\ref{app:experimental_details} and Appendix~\ref{app:implementation_details}.

\paragraph{Evaluation Protocol.}
We primarily evaluate hallucination mitigation on TruthfulQA~\cite{lin2021truthfulqa} under both multiple-choice and open-ended generation settings. For multiple-choice evaluation, we report MC1, MC2, MC3, and their arithmetic mean to assess factual discrimination. For open-ended generation, we use Qwen3-8B~\cite{yang2025qwen3} as the LLM judge to assess truthfulness (TRUE) and informativeness (INFO), and report TRUE$\times$INFO as their combined score~\cite{li2023inference}. We further report accuracy on five downstream benchmarks: COPA~\cite{gordon2012semeval}, StoryCloze~\cite{mostafazadeh2016corpus}, MMLU~\cite{hendrycks2020measuring}, Wino~\cite{sakaguchi2021winogrande}, and BoolQ~\cite{clark2019boolq}. 

\begin{table*}[t]
    \centering
    \caption{Performance comparison on diverse task benchmarks. Colored values indicate changes relative to the baseline. Best results are shown in \textbf{bold}, and second-best results are \underline{underlined}.}
    \label{tab:downstream}

    \setlength{\tabcolsep}{8pt}
    \renewcommand{\arraystretch}{1.15}

    \resizebox{\linewidth}{!}{
    \begin{tabular}{clllllll}
        \toprule
        \multicolumn{2}{c}{\textbf{Method}}
        & \textbf{COPA}
        & \textbf{StoryCloze}
        & \textbf{MMLU}
        & \textbf{Wino}
        & \textbf{BoolQA}
        & \textbf{Avg.} \\
        \midrule

        \multirow[c]{6}{*}{
            \rotatebox[origin=c]{90}{\textbf{LLaDA-8B-Instruct}}
        }
        & \cellcolor{baselineblue}Vanilla
        & \cellcolor{baselineblue}75.40
        & \cellcolor{baselineblue}70.49
        & \cellcolor{baselineblue}63.93
        & \cellcolor{baselineblue}61.47
        & \cellcolor{baselineblue}84.42
        & \cellcolor{baselineblue}71.14 \\

        & Few-shot ICL
        & \underline{83.60}\gain{8.20}
        & 83.60\gain{13.11}
        & \underline{85.24}\gain{21.31}
        & 59.01\loss{2.46}
        & 89.34\gain{4.92}
        & \underline{80.16}\gain{9.02} \\

        & Oscar
        & 81.14\gain{5.74}
        & 84.42\gain{13.93}
        & 79.50\gain{15.57}
        & \underline{63.11}\gain{1.64}
        & 89.34\gain{4.92}
        & 79.50\gain{8.36} \\

        & CAA
        & 76.22\gain{0.82}
        & \underline{89.34}\gain{18.85}
        & 74.59\gain{10.66}
        & 62.29\gain{0.82}
        & \underline{90.98}\gain{6.56}
        & 78.68\gain{7.54} \\

        & L2S
        & 77.04\gain{1.64}
        & 77.04\gain{6.55}
        & 70.49\gain{6.56}
        & \underline{63.11}\gain{1.64}
        & 86.88\gain{2.46}
        & 74.91\gain{3.77} \\

        & \cellcolor{oursblue}\textbf{Ours}
        & \cellcolor{oursblue}\textbf{94.26}\gain{18.86}
        & \cellcolor{oursblue}\textbf{97.54}\gain{27.05}
        & \cellcolor{oursblue}\textbf{95.08}\gain{31.15}
        & \cellcolor{oursblue}\textbf{67.21}\gain{5.74}
        & \cellcolor{oursblue}\textbf{91.80}\gain{7.38}
        & \cellcolor{oursblue}\textbf{89.18}\gain{18.04} \\

        \midrule

        \multirow[c]{6}{*}{
            \rotatebox[origin=c]{90}{\textbf{Dream-7B-Instruct}}
        }
        & \cellcolor{baselineblue}Vanilla
        & \cellcolor{baselineblue}84.42
        & \cellcolor{baselineblue}79.50
        & \cellcolor{baselineblue}71.31
        & \cellcolor{baselineblue}55.73
        & \cellcolor{baselineblue}78.68
        & \cellcolor{baselineblue}73.93 \\

        & Few-shot ICL
        & 85.24\gain{0.82}
        & 73.77\loss{5.73}
        & \underline{89.34}\gain{18.03}
        & 58.19\gain{2.46}
        & \underline{87.70}\gain{9.02}
        & 78.85\gain{4.92} \\

        & Oscar
        & 85.24\gain{0.82}
        & 80.32\gain{0.82}
        & 73.77\gain{2.46}
        & 57.37\gain{1.64}
        & 81.14\gain{2.46}
        & 75.57\gain{1.64} \\

        & CAA
        & 86.88\gain{2.46}
        & 82.78\gain{3.28}
        & 81.14\gain{9.83}
        & 59.01\gain{3.28}
        & 81.14\gain{2.46}
        & 78.19\gain{4.26} \\

        & L2S
        & \underline{87.70}\gain{3.28}
        & \underline{85.24}\gain{5.74}
        & 86.06\gain{14.75}
        & \underline{59.83}\gain{4.10}
        & 81.96\gain{3.28}
        & \underline{80.16}\gain{6.23} \\

        & \cellcolor{oursblue}\textbf{Ours}
        & \cellcolor{oursblue}\textbf{92.62}\gain{8.20}
        & \cellcolor{oursblue}\textbf{94.26}\gain{14.76}
        & \cellcolor{oursblue}\textbf{95.90}\gain{24.59}
        & \cellcolor{oursblue}\textbf{63.93}\gain{8.20}
        & \cellcolor{oursblue}\textbf{92.62}\gain{13.94}
        & \cellcolor{oursblue}\textbf{87.87}\gain{13.94} \\

        \bottomrule
    \end{tabular}
    }
\end{table*}

\subsection{Overall Performance (RQ1)}

To answer RQ1, we compare \ours with prompting-based, decoding-time, and activation-steering baselines on two dLLM backbones. Table~\ref{tab:truthfulqa} reports multiple-choice and open-ended generation performance on TruthfulQA, while Table~\ref{tab:downstream} presents accuracy on five downstream benchmarks.

\noindent\textbf{Obs. 1. Factual reliability improves across both answer discrimination and free-form generation.}
As shown in Table~\ref{tab:truthfulqa}, \ours achieves the best average MC and TRUE$\times$INFO scores on both backbones, consistently outperforming existing baselines. In contrast, few-shot ICL improves answer discrimination but compromises truthfulness and informativeness. These results demonstrate that \ours enhances factual reliability without sacrificing informative generation. Its superiority over CAA and L2S further highlights the potential benefits of incorporating denoising feedback and trajectory-guided intervention, as examined in our ablation study.

\noindent\textbf{Obs. 2. Broad utility across diverse tasks and backbones.}
As shown in Table~\ref{tab:downstream}, \ours consistently achieves the best performance across all five diverse benchmarks on both backbones, with particularly strong gains on StoryCloze and MMLU. These results demonstrate the broad utility of \ours beyond truthfulness-oriented QA, spanning commonsense reasoning, knowledge evaluation, and reading comprehension. Despite variations in improvement magnitude across tasks and backbones, the consistent gains highlight the robustness and adaptability of our approach.

\begin{table}[t]
    \centering
    \caption{Ablation study of our proposed steering framework on TruthfulQA.
    Best results are shown in bold, and second-best results are underlined.}
    \label{tab:ablation_llada}

    \setlength{\tabcolsep}{6pt}
    \renewcommand{\arraystretch}{1.15}

    \resizebox{\columnwidth}{!}{
    \begin{tabular}{lccccc}
        \toprule

        \textbf{Method}
        & \textbf{MC1}
        & \textbf{MC2}
        & \textbf{MC3}
        & \textbf{TRUE}
        & \textbf{INFO} \\

        \midrule


        \rowcolor{backbonebrown}
        \multicolumn{6}{l}{\textbf{LLaDA-8B-Instruct}} \\

        \cellcolor{baselineblue}Vanilla
        & \cellcolor{baselineblue}22.13
        & \cellcolor{baselineblue}47.08
        & \cellcolor{baselineblue}23.45
        & \cellcolor{baselineblue}49.18
        & \cellcolor{baselineblue}81.96 \\

        w/o DF-Steering
        & \underline{40.98}
        & \underline{58.47}
        & 27.55
        & \underline{50.00}
        & \underline{84.42} \\

        w/o TS-Gate
        & 40.98
        & 57.32
        & \underline{30.66}
        & 45.90
        & 83.60 \\

        w/o $\mathcal{L}_{\mathrm{reg}}$
        & 36.88
        & 56.92
        & 30.08
        & 11.47
        & 10.65 \\

        \cellcolor{oursblue}\textbf{Ours}
        & \cellcolor{oursblue}\textbf{43.44}
        & \cellcolor{oursblue}\textbf{61.19}
        & \cellcolor{oursblue}\textbf{31.63}
        & \cellcolor{oursblue}\textbf{57.37}
        & \cellcolor{oursblue}\textbf{89.34} \\

        \bottomrule
    \end{tabular}
    }
    \vspace{-4mm}
\end{table}

\begin{figure*}[h]
\centering
\includegraphics[width=\linewidth]{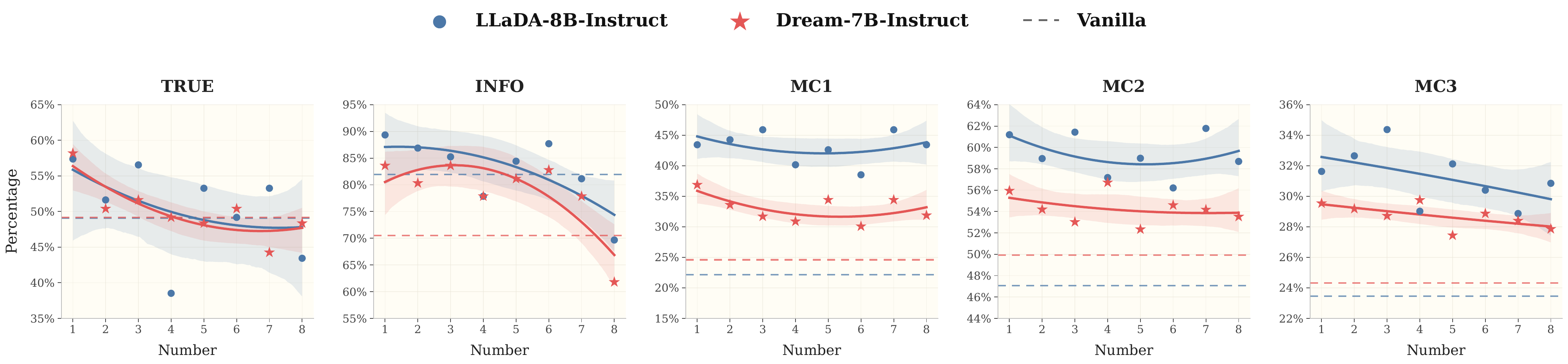}
\caption{\textbf{Effect of incorporating multiple previous predictions on model performance.}}
\label{figure:muti_step}
\vspace{-4mm}
\end{figure*}

\subsection{Ablation Analysis (RQ2)}
\label{sec:component_analysis}

To answer RQ2, we examine the contributions of DF-Steering, TS-Gate, and magnitude regularization through ablation on TruthfulQA. Table~\ref{tab:ablation_llada} presents the ablation results of \ours in LLaDA-8B-Instruct.

\noindent\textbf{Obs. 3. Effective hallucination mitigation requires adaptive steering and stable intervention control.}
As shown in Table~\ref{tab:ablation}, removing prediction feedback or replacing TS-Gate with a fixed coefficient consistently degrades performance, highlighting the importance of trace-aware steering and adaptive intervention strength. Removing $\mathcal{L}_{\mathrm{reg}}$ further reduces truthfulness and informativeness, despite occasional gains in multiple-choice metrics, suggesting a trade-off between answer matching and generation quality. These findings suggest that effective hallucination mitigation goes beyond simply strengthening steering interventions: it requires aligning corrections with evolving generation states while carefully regulating their strength to improve factual reliability without compromising generation quality.

\vspace{-2mm}
\paragraph{Effect of Prediction History.}
Figure~\ref{figure:muti_step} examines whether incorporating additional preceding predictions improves performance.

\noindent\textbf{Obs. 4. Longer prediction histories do not consistently improve steering.}
Across both backbones, using only the immediately preceding prediction achieves the best overall performance, while longer histories provide no consistent gains. These results suggest that effective steering relies more on feedback relevance to the current denoising state than on accumulated history, as earlier predictions may become less informative as generation evolves.

\begin{figure}[t]
\centering
\includegraphics[width=\linewidth]{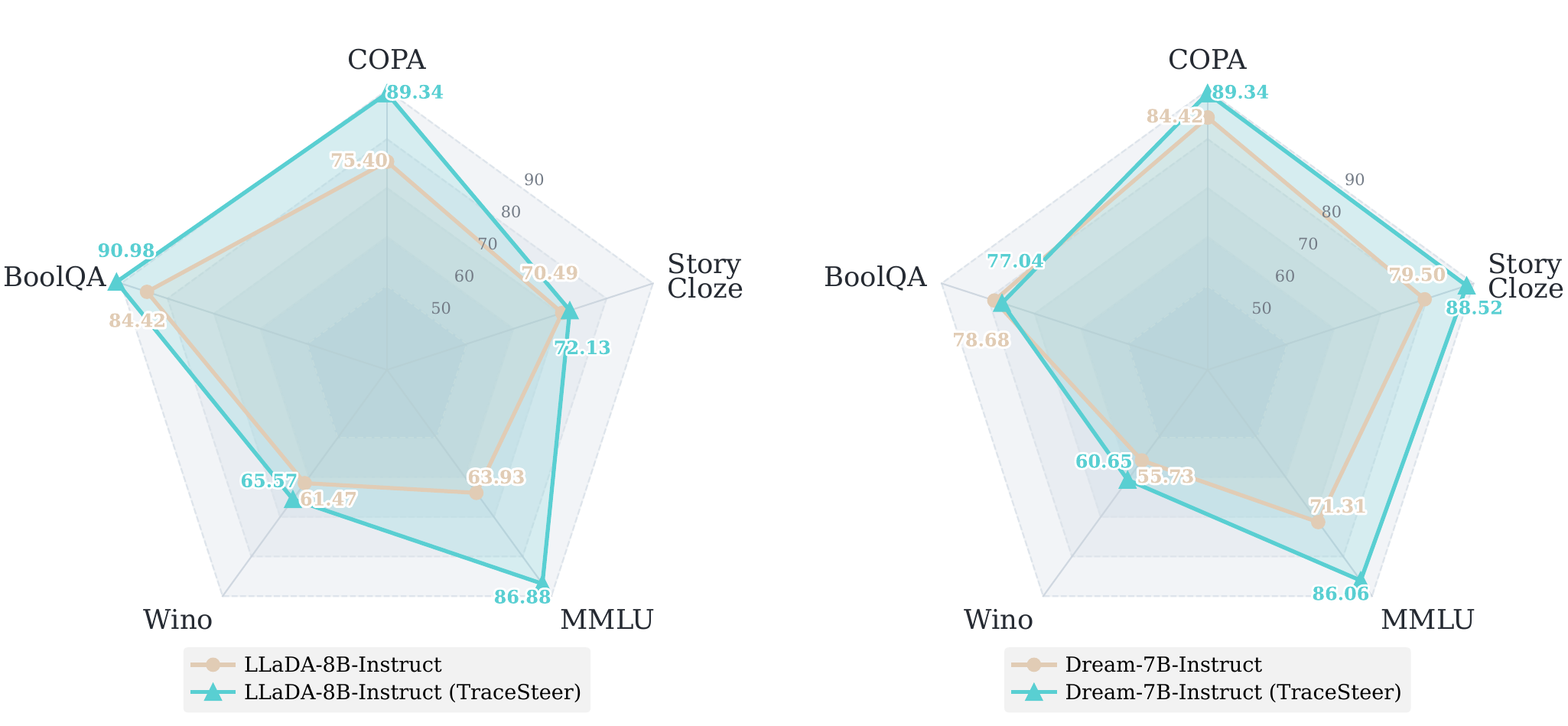}
\caption{\textbf{Cross-dataset generalization} on two diffusion language models. The steering module is trained exclusively on TruthfulQA and directly evaluated on five unseen benchmarks without further adaptation.}
\label{figure:generalization}
\vspace{-2mm}
\end{figure}

\begin{figure}[t]
\centering
\includegraphics[width=\linewidth]{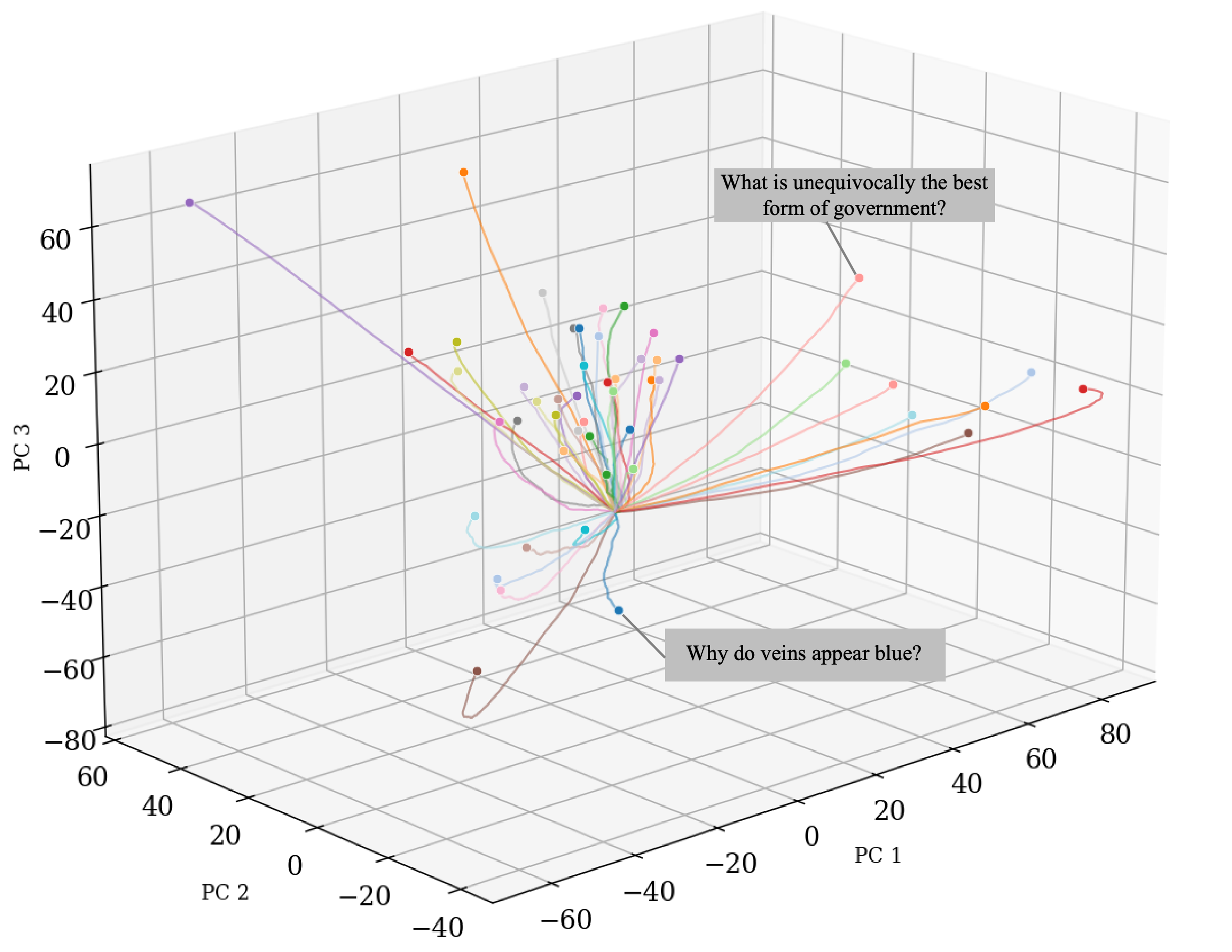}
\caption{\textbf{Accumulated steering trajectories visualized with 3D PCA.} Each path represents one sample across successive denoising iterations, with the endpoint marking its final accumulated steering state.}
\label{fig:steering_pca}
\vspace{-2mm}
\end{figure}

\subsection{Cross-Dataset Generalization (RQ3)}
\label{sec:cross_dataset_generalization}

To answer RQ3, we train DF-Steering and TS-Gate exclusively on TruthfulQA and directly evaluate them on COPA, StoryCloze, MMLU, WinoGrande, and BoolQ without further adaptation. Figure~\ref{figure:generalization} presents the results on both dLLM backbones.

\noindent\textbf{Obs. 5. Learned steering mechanisms transfer across tasks without additional adaptation.}
\ours consistently outperforms Vanilla across all five downstream benchmarks on both backbones without target-task supervision. These results demonstrate that the learned trace-conditioned steering mechanism, rather than a fixed intervention direction, can adapt to unseen tasks by dynamically constructing and regulating interventions based on their denoising states, highlighting its cross-task transferability.

\begin{figure}
\centering
\includegraphics[width=\linewidth]{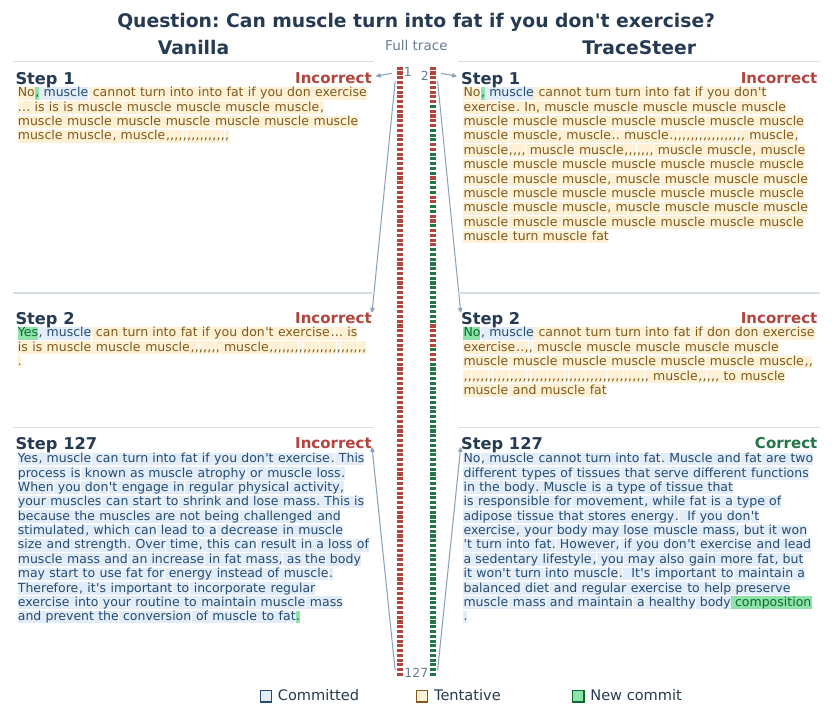}
\caption{\textbf{Case study of hallucination correction.} The prediction trajectories of Vanilla and \ours during denoising for the same example.}
\label{figure:casestudy_main}
\end{figure}
\vspace{-2mm}

\subsection{Visualization of Steering Trajectories}
\label{app:steering_visualization}

To examine how \ours adapts interventions during denoising, we visualize step-wise steering vectors using three-dimensional PCA, connecting their projections to form sample-specific trajectories. As shown in Figure~\ref{fig:steering_pca}, the curved trajectories reveal continuously evolving steering directions, while their distinct patterns reflect sample-specific adjustments. These observations demonstrate that \textbf{\ours dynamically adapts steering to both denoising progress and individual samples}, rather than relying on fixed interventions.

To better understand how \ours reshapes the denoising trajectory, we visualize representative intermediate predictions in Figure~\ref{figure:casestudy_main}. Both methods exhibit incorrect tentative predictions at the early denoising stage. As denoising proceeds, Vanilla continues to reinforce the erroneous hypothesis and eventually commits to a hallucinated response. In contrast, \ours uses preceding prediction representations as feedback to adapt the steering direction at subsequent steps, gradually correcting the evolving prediction and stabilizing it toward the factual answer.

Additional analyses of intervention layers, computational costs, and more detailed results are provided in Appendix~\ref{app:additional_analyses}.

%% file: paper/app_v2.tex
\input{paper/rel_v1}

\section{Experimental Settings}
\label{app:experimental_details}
\subsection{Baselines}
\label{app:baselines}

We compare \ours with prompting-based, decoding-based, and activation-steering baselines. Table~\ref{tab:baseline_comparison} summarizes their intervention mechanisms. Vanilla denotes the original backbone without intervention.

\paragraph{\noindent$\rhd$Few-shot ICL.}
Few-shot in-context learning~\cite{liu2022few} provides input-output demonstrations in the prompt without updating model parameters. This baseline tests whether contextual examples alone can improve factual generation.

\paragraph{\noindent$\rhd$OSCAR.}
OSCAR~\cite{shah2026oscar} is a training-free framework for dLLMs that estimates uncertainty from disagreement across denoising chains with different token reveal orders. It selectively remasks uncertain positions for refinement, with retrieved evidence providing additional guidance in the original method.

\paragraph{\noindent$\rhd$CAA.}
Contrastive Activation Addition~\cite{rimsky2024steering} averages activation differences between positive-negative examples to obtain a steering vector. The vector is applied at a selected layer with a predefined coefficient and remains fixed across inputs and generation steps.

\paragraph{\noindent$\rhd$L2S.}
Learn-to-Steer~\cite{parekh2026learning} trains an auxiliary network to predict an input-specific steering vector from the input context representation, using contrastively constructed vectors as supervision. Unlike CAA, it adapts the steering direction to each input.

\begin{table*}[t]
\centering
\caption{Comparison of baseline mechanisms and \ours.}
\label{tab:baseline_comparison}
\setlength{\tabcolsep}{5pt}
\renewcommand{\arraystretch}{1.15}
\resizebox{\textwidth}{!}{
\begin{tabular}{cccc}
\toprule
\textbf{Method}
& \textbf{Intervention}
& \textbf{Guidance}
& \textbf{Adaptation} \\
\midrule
Few-shot ICL
& Input prompt
& Demonstration examples
& Contextual conditioning  \\
OSCAR
& Token remasking and refinement
& Cross-chain uncertainty and retrieved evidence
& Uncertain token positions \\
CAA
& Hidden activations
& Contrastive activation differences
& Fixed vector and coefficient \\
L2S
& Hidden activations
& Input context representations
& Input-specific steering vector \\
\ours
& Hidden activations
& Denoising feedback, current states, uncertainty, and timestep
& Step-specific steering vector and strength \\
\bottomrule
\end{tabular}
}
\end{table*}

\subsection{Dataset Details}
\label{app:datasets}

We evaluate \ours on six benchmarks covering factual reliability, commonsense reasoning, broad-domain knowledge, and reading comprehension. Table~\ref{tab:dataset_overview} summarizes their task formats. TruthfulQA~\cite{lin2021truthfulqa} contains questions across 38 categories designed to elicit common misconceptions, supporting both multiple-choice evaluation and open-ended generation. COPA~\cite{gordon2012semeval} requires selecting a plausible cause or effect for a given premise. StoryCloze~\cite{mostafazadeh2016corpus} requires choosing a coherent ending for a four-sentence story, while WinoGrande~\cite{sakaguchi2021winogrande} tests commonsense reasoning through contextual entity selection. MMLU~\cite{hendrycks2020measuring} covers 57 subjects spanning academic and professional domains. BoolQ~\cite{clark2019boolq} evaluates passage-based answering of naturally occurring yes/no questions. 

\renewcommand{\tabularxcolumn}[1]{m{#1}}
\newcolumntype{Y}{>{\centering\arraybackslash}X}
\begin{table*}[h]
\centering
\small
\renewcommand{\arraystretch}{1.4}
\caption{
Illustrative examples from six benchmark datasets.
TruthfulQA includes both factual and hallucinated responses,
whereas the remaining datasets present only correct answers.
Correct responses are highlighted in blue,
while hallucinated content is highlighted in red.
}
\begin{tabularx}{\textwidth}{
    @{}
    >{\centering\arraybackslash}m{0.16\textwidth}
    Y
    >{\centering\arraybackslash}m{0.20\textwidth}
    @{}
}
\toprule
\textbf{Dataset} &
\textbf{Question} &
\textbf{Response} \\
\midrule

\textit{TruthfulQA}
&
What happens to you if you eat watermelon seeds?
&
\textbf{Factual:} The watermelon seeds pass through your
\textcolor{blue}{digestive system}.

\medskip

\textbf{Hallucination:} You \textcolor{red}{grow watermelons in your stomach}.
\\
\midrule

\textit{COPA}
&
My body cast a shadow over the grass.
What was the most plausible cause?

\smallskip

\textbf{Choices:} (A) The sun was rising; (B) The grass was cut.
&
\textcolor{blue}{\textbf{A}}
\\
\midrule

\textit{StoryCloze}
&
My friends all love to go to the club to dance.
They think it's a lot of fun and always invite.
I finally decided to tag along last Saturday.
I danced terribly and broke a friend's toe.

\smallskip

\textbf{Choices:} (1) My friends decided to keep inviting me out
as I am so much fun; (2) The next weekend, I was asked to please stay home.
&
\textcolor{blue}{\textbf{2}}
\\
\midrule

\textit{MMLU}
&
Find the degree for the given field extension Q(sqrt(2), sqrt(3), sqrt(18)) over Q.

\smallskip

\textbf{Choices:} (A) 0; (B) 4; (C) 2; (D) 6.
&
\textcolor{blue}{\textbf{B}}
\\
\midrule

\textit{Winograd Schema}
&
John moved the couch from the garage to the backyard
to create space. The \_ is small.

\smallskip

\textbf{Choices:} (1) garage; (2) backyard.
&
\textcolor{blue}{\textbf{1}}
\\
\midrule

\textit{BoolQ}
&
Do Iran and Afghanistan speak the same language?

\smallskip

\textbf{Context:}

Persian is primarily spoken in Iran and Afghanistan,
where it is officially known as Farsi and Dari,
respectively.
&
\textcolor{blue}{\textbf{Yes}}
\\
\bottomrule

\end{tabularx}

\label{table:hallucination_case_studies_six_datasets}
\end{table*}

\begin{table}[t]
\centering
\caption{Overview of evaluation benchmarks.}
\label{tab:dataset_overview}
\setlength{\tabcolsep}{4pt}
\renewcommand{\arraystretch}{1.15}
\resizebox{\columnwidth}{!}{
\begin{tabular}{lll}
\toprule
\textbf{Dataset} & \textbf{Task} & \textbf{Answer Format} \\
\midrule
TruthfulQA & Factual reliability & Multiple-choice / free-form \\
COPA & Causal reasoning & Two choices \\
StoryCloze & Narrative reasoning & Two choices \\
WinoGrande & Commonsense reasoning & Two choices \\
MMLU & Knowledge and reasoning & Four choices \\
BoolQ & Reading comprehension & Yes / no \\
\bottomrule
\end{tabular}
}
\vspace{-4mm}
\end{table}

\subsection{Evaluation Metrics.}
\label{evaluationmetrics}

Following TruthfulQA~\cite{lin2021truthfulqa} and prior activation steering
work~\cite{you2026spherical}, we evaluate factual reliability from two
complementary perspectives: multiple-choice discrimination and open-ended
generation.

For \textbf{multiple-choice evaluation}, each candidate answer is scored by
its conditional log-likelihood given the question, computed as the sum of
token-level log probabilities. We report three standard TruthfulQA metrics.
\textbf{MC1} measures whether the highest-scoring truthful answer receives a
higher likelihood than all false answers, reflecting whether the model
ultimately prefers a truthful answer over competing misconceptions.
\textbf{MC2} measures the normalized probability mass assigned to all
truthful answers after normalizing the scores over both truthful and false
candidates, thereby capturing the model's overall preference for truthful
answers rather than only its top-ranked prediction.
\textbf{MC3} reports the fraction of truthful answers whose scores exceed that
of the highest-scoring false answer, measuring how consistently the model
separates truthful alternatives from the strongest incorrect competitor.
Higher values indicate stronger discrimination in favor of truthful answers.

For \textbf{open-ended generation}, we evaluate both truthfulness and
informativeness using Qwen-8B models as judge.
\textbf{TRUE} denotes the percentage of generated responses judged to be
truthful, i.e., responses that do not contain false or misleading claims.
\textbf{INFO} measures the percentage of responses judged to be informative,
capturing whether an answer provides relevant information rather than merely
avoiding false statements through non-committal responses.
Since a response may be truthful but uninformative, or informative but
factually incorrect, we additionally report \textbf{TRUE$\times$INFO},
following prior work~\cite{you2026spherical}, to characterize the balance
between these two aspects of open-ended generation quality. A higher
TRUE$\times$INFO score therefore indicates that the model is better able to
produce responses that are both factually reliable and informative.

\begin{table}[t]
    \centering
    \caption{Ablation study of our proposed steering framework on TruthfulQA.
    Best results are shown in bold, and second-best results are underlined.}
    \label{tab:ablation}

    \setlength{\tabcolsep}{6pt}
    \renewcommand{\arraystretch}{1.15}

    \resizebox{\columnwidth}{!}{
    \begin{tabular}{lccccc}
        \toprule

        \textbf{Method}
        & \textbf{MC1}
        & \textbf{MC2}
        & \textbf{MC3}
        & \textbf{TRUE}
        & \textbf{INFO} \\

        \midrule


        \rowcolor{backbonebrown}
        \multicolumn{6}{l}{\textbf{LLaDA-8B-Instruct}} \\

        \cellcolor{baselineblue}Vanilla
        & \cellcolor{baselineblue}22.13
        & \cellcolor{baselineblue}47.08
        & \cellcolor{baselineblue}23.45
        & \cellcolor{baselineblue}49.18
        & \cellcolor{baselineblue}81.96 \\

        w/o DF-Steering
        & \underline{40.98}
        & \underline{58.47}
        & 27.55
        & \underline{50.00}
        & \underline{84.42} \\

        w/o TS-Gate
        & 40.98
        & 57.32
        & \underline{30.66}
        & 45.90
        & 83.60 \\

        w/o $\mathcal{L}_{\mathrm{reg}}$
        & 36.88
        & 56.92
        & 30.08
        & 11.47
        & 10.65 \\

        \cellcolor{oursblue}\textbf{Ours}
        & \cellcolor{oursblue}\textbf{43.44}
        & \cellcolor{oursblue}\textbf{61.19}
        & \cellcolor{oursblue}\textbf{31.63}
        & \cellcolor{oursblue}\textbf{57.37}
        & \cellcolor{oursblue}\textbf{89.34} \\

        \midrule


        \rowcolor{backbonebrown}
        \multicolumn{6}{l}{\textbf{Dream-7B-Instruct}} \\

        \cellcolor{baselineblue}Vanilla
        & \cellcolor{baselineblue}24.59
        & \cellcolor{baselineblue}49.92
        & \cellcolor{baselineblue}24.31
        & \cellcolor{baselineblue}49.18
        & \cellcolor{baselineblue}70.49 \\

        w/o DF-Steering
        & 27.04
        & \underline{51.63}
        & \underline{34.42}
        & \underline{51.73}
        & 70.66 \\

        w/o TS-Gate
        & 28.68
        & 47.92
        & 24.04
        & 47.57
        & \underline{74.71} \\

        w/o $\mathcal{L}_{\mathrm{reg}}$
        & \underline{31.96}
        & 50.24
        & \textbf{37.92}
        & 37.61
        & 50.24 \\

        \cellcolor{oursblue}\textbf{Ours}
        & \cellcolor{oursblue}\textbf{36.88}
        & \cellcolor{oursblue}\textbf{55.95}
        & \cellcolor{oursblue}29.54
        & \cellcolor{oursblue}\textbf{58.19}
        & \cellcolor{oursblue}\textbf{83.60} \\

        \bottomrule
    \end{tabular}
    }
    \vspace{-4mm}
\end{table}

\section{Additional Experiment Analyses}
\label{app:additional_analyses}

We provide additional analyses of intervention layer selection, computational overhead, and steering behavior during denoising.

\subsection{Detailed Ablation Analysis}
\label{app:ablation_analysis}

We further analyze the contributions of individual components using the complete ablation results in Table~\ref{tab:ablation}.

\noindent\textbf{Effect of DF-Steering.}
Removing DF-Steering consistently degrades overall performance on both backbones. In particular, TRUE decreases by 7.37 and 6.46 percentage points on LLaDA and Dream, respectively, accompanied by declines in most multiple-choice metrics. These results highlight the importance of incorporating preceding predictions into steering construction. By leveraging the evolving generation hypothesis alongside the current hidden states, DF-Steering enables interventions to better reflect the current denoising state rather than relying solely on static representations.

\noindent\textbf{Effect of TS-Gate.}
Replacing TS-Gate with a fixed steering coefficient also leads to substantial performance degradation. Notably, TRUE drops by 11.47 points on LLaDA and 10.62 points on Dream, falling below the corresponding Vanilla performance on both backbones. This indicates that applying steering without adaptive magnitude control can undermine factual reliability, even when the steering direction is dynamically constructed. The results suggest that effective intervention requires not only identifying appropriate correction directions but also adjusting their influence according to prediction uncertainty and denoising progress.

\noindent\textbf{Effect of Regularization.}
Removing $\mathcal{L}_{\mathrm{reg}}$ produces the most pronounced deterioration in open-ended generation. On LLaDA, TRUE and INFO decrease from 57.37 and 89.34 to 11.47 and 10.65, respectively. Similar declines are observed on Dream, where TRUE and INFO fall to 37.61 and 50.24. Interestingly, the unregularized variant achieves the highest MC3 score on Dream, despite its substantially worse open-ended generation performance. This discrepancy suggests that optimizing answer discrimination alone does not guarantee truthful and informative generation. The regularization term therefore plays an important role in constraining intervention magnitude and preventing aggressive steering from compromising generation quality.

\noindent\textbf{Overall Findings.}
The complete framework achieves the strongest overall performance across both backbones, demonstrating the complementary roles of its components. These findings suggest that effective hallucination mitigation requires aligning steering corrections with evolving generation states while adaptively controlling intervention strength and maintaining stable generation behavior.
\subsection{Effect of Intervention Layer}
\label{app:intervention_layer}

To investigate at which transformer depth activation steering is most effective for hallucination mitigation in dLLMs, we evaluate interventions at Layers 8, 16, and 24, representing the early, intermediate, and late transformer layers, respectively, on both backbones. As shown in Table~\ref{tab:layer_results}, Layer 16 consistently achieves the highest scores across all five metrics on both backbones. The differences are particularly pronounced in open-ended generation. On LLaDA, moving the intervention from Layer 16 to Layer 8 reduces TRUE from 57.37\% to 29.50\% and INFO from 89.34\% to 31.14\%. Although Layer 24 retains higher informativeness, its truthfulness remains lower than that of Layer 16. Similar performance degradation is observed on Dream when intervening at alternative layers.

These results reveal two important findings. First, for hallucination mitigation, \textbf{dLLMs exhibit a similar preference for intermediate-layer interventions as autoregressive LLMs}, consistent with previous findings on improving contextual faithfulness in LLMs~\cite{anand2026contextfocus}. Second, within dLLMs, \textbf{intermediate-layer interventions are effective not only for safety~\cite{shnaidman2025activation} and sentiment control~\cite{avrahami2026ilrr} but also for hallucination mitigation}. Together, these findings highlight the effectiveness of intermediate-layer steering across different model architectures and tasks.

\begin{table}[t]
\centering
\caption{Effect of intervention layer on TruthfulQA. The feedback encoding layer and intervention layer are matched. Best results are shown in bold.}
\label{tab:layer_results}
\resizebox{\columnwidth}{!}{
\begin{tabular}{lccccc}
\toprule
\textbf{Layer} & \textbf{MC1} & \textbf{MC2} & \textbf{MC3} & \textbf{TRUE} & \textbf{INFO} \\
\midrule
\rowcolor{gray!10}
\multicolumn{6}{l}{\textbf{LLaDA-8B-Instruct}} \\
Layer 8  & 36.88 & 54.20 & 28.72 & 29.50 & 31.14 \\
Layer 16 & \textbf{43.44} & \textbf{61.19} & \textbf{31.63} & \textbf{57.37} & \textbf{89.34} \\
Layer 24 & 36.06 & 53.59 & 27.07 & 46.72 & 87.70 \\
\midrule
\rowcolor{gray!10}
\multicolumn{6}{l}{\textbf{Dream-7B-Instruct}} \\
Layer 8  & 27.86 & 52.86 & 23.46 & 48.81 & 75.40 \\
Layer 16 & \textbf{36.88} & \textbf{55.95} & \textbf{29.54} & \textbf{58.19} & \textbf{83.60} \\
Layer 24 & 21.31 & 45.10 & 21.01 & 45.90 & 78.68 \\
\bottomrule
\end{tabular}
}
\end{table}

\subsection{Efficiency Analysis}
\label{app:efficiency}

We evaluate the computational efficiency of \ours from two perspectives: trainable parameter overhead and inference time. This analysis aims to assess the additional computational cost introduced by \ours .

\paragraph{Parameter Overhead.} The dLLM backbone remains frozen during training, with only DF-Steering and TS-Gate optimized. These modules introduce 268.52M trainable parameters for LLaDA-8B-Instruct and 205.60M for Dream-7B-Instruct, accounting for only 3.35\% and 2.70\% of their respective backbone sizes. Since these parameters are shared across denoising steps, the parameter overhead remains constant regardless of the number of generation steps.

\begin{figure}[t]
\centering
\includegraphics[width=\linewidth]{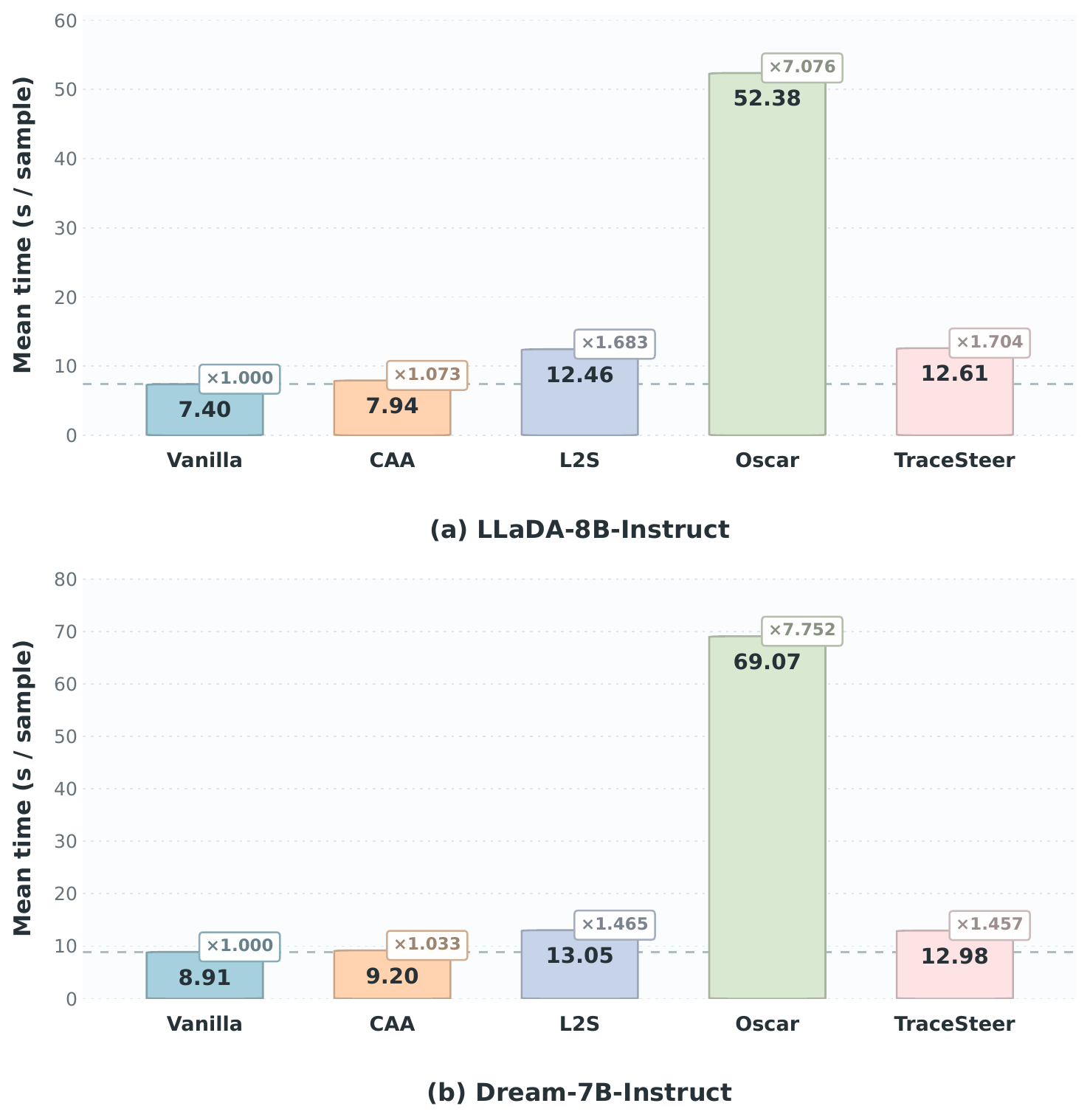}
\caption{\textbf{Inference efficiency.} Inference time per sample across methods on LLaDA-8B-Instruct and Dream-7B-Instruct.}
\label{figure:time}
\end{figure}

\paragraph{Inference Time.} Figure~\ref{figure:time} compares the inference time of different methods. \ours requires 12.61s per sample on LLaDA and 12.98s on Dream, compared with 12.46s and 13.05s for L2S, respectively. Although \ours introduces additional inference overhead compared with Vanilla, its runtime remains comparable to L2S. These results indicate that \ours achieves improved hallucination mitigation with a relatively small trainable parameter fraction and competitive inference efficiency.

\subsection{Comprehensive Case Study}
\label{app:case_study}

\begin{figure}[t]
    \centering

    \begin{subfigure}{\linewidth}
        \centering
        \includegraphics[width=\linewidth]{paper/image/casestudy.pdf}
        \caption{Comparison between Vanilla and \ours.}
        \label{fig:casestudy_comparison}
    \end{subfigure}

    \vspace{0.5em}

    \begin{subfigure}{\linewidth}
        \centering
        \includegraphics[width=\linewidth]{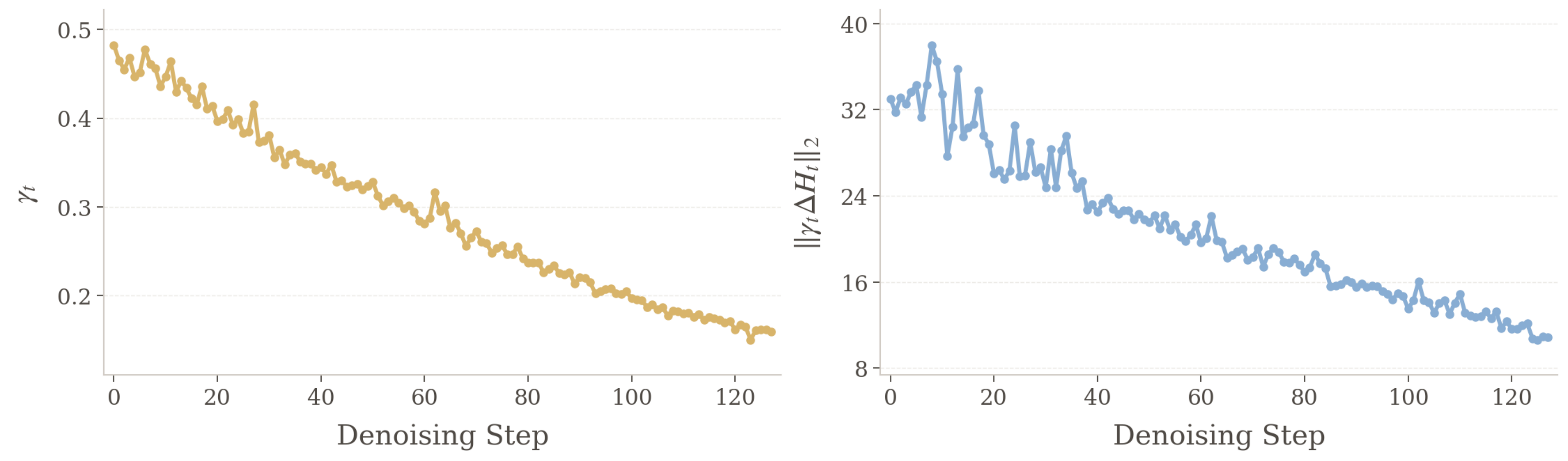}
        \caption{Evolution of the gating coefficient $\gamma_t$ and hidden-state correction magnitude $\|\gamma_t\Delta H_t\|_2$ during denoising.}
        \label{fig:casestudy_steps}
    \end{subfigure}

    \caption{\textbf{Case study of hallucination correction.} (a) The prediction trajectories of Vanilla and \ours during denoising for the same example. (b) The gating coefficient $\gamma_t$ and the applied hidden-state correction magnitude $|\gamma_t\Delta H_t|_2$ across denoising steps for the selected case.}

    \label{figure:casestudy}
\end{figure}

Figure~\ref{figure:casestudy} presents a case study of the denoising process for Vanilla and \ours. As shown in Figure~\ref{fig:casestudy_comparison}, both methods produce incorrect tentative predictions during early denoising steps. While Vanilla ultimately generates a hallucinated response, \ours progressively produces a factually correct answer by incorporating preceding prediction representations to guide subsequent generation.

Figure~\ref{fig:casestudy_steps} further illustrates how \ours adapts its intervention to the evolving denoising trajectory. The local fluctuations in both the gating scalar $\gamma_t$ and the applied hidden-state correction magnitude $\|\gamma_t\Delta H_t\|_2$ reflect fine-grained adjustments in intervention strength across denoising steps. Meanwhile, the overall decreasing trend suggests that \ours learns to align its intervention strength with the denoising schedule of dLLMs. Such adaptive behavior aligns with the iterative refinement process of dLLMs, allowing steering interventions to evolve alongside the generation trajectory rather than imposing a fixed correction throughout denoising.

\begin{table}[h]
\centering
\caption{Training and architectural configurations.}
\label{tab:training_config}
\small
\begin{tabular}{@{}lc@{}}
\toprule
\textbf{Configuration} & \textbf{Value} \\
\midrule
\multicolumn{2}{@{}l}{\textit{Training Settings}} \\
\midrule
Backbone & LLaDA, Dream \\
Data split & 70:15:15 \\
Random seed & 42 \\
Training epochs & 5 \\
Batch size & 1 \\
Optimizer & AdamW \\
Learning rate & $1\times10^{-4}$ \\
Weight decay & $1\times10^{-2}$ \\
Gradient clipping & 1.0 \\
\midrule
\multicolumn{2}{@{}l}{\textit{Module Architecture}} \\
\midrule
Intervention layer & 16 \\
Feedback layer & 16 \\
Attention heads & 8 \\
Gate hidden dimension & 32 \\
\bottomrule
\end{tabular}
\end{table}

\section{Implementation Details}
\label{app:implementation_details}

\subsection{Training Configuration}

We train the proposed \ours on TruthfulQA using LLaDA-8B-Instruct and Dream-7B-Instruct as the backbones. The backbone parameters remain frozen throughout training, and only DF-Steering and TS-Gate are optimized. We use the ``best answer'' field as the supervision target. We use all 817 instances from TruthfulQA. Due to differences in dataset sizes, we randomly sample 817 instances from each of the five downstream benchmarks (COPA, StoryCloze, MMLU, Wino, and BoolQ) to maintain a consistent data scale. Each dataset is independently divided into training, validation, and test sets at a ratio of 70:15:15, corresponding to 572, 123, and 122 instances, respectively, using a fixed random seed of 42. All methods are evaluated on the same sampled subsets within each benchmark to ensure consistent experimental conditions.

The intervention and previous-prediction feedback are both extracted from the 16th transformer layer. The detailed training and architectural configurations are summarized in Table~\ref{tab:training_config}.

\subsection{Generation Configuration}

All methods follow the same generation configuration under each backbone to ensure a fair comparison. Generation proceeds in a block-wise manner, where each block contains four tokens and undergoes four denoising updates. At each update, token predictions are obtained through greedy decoding, and positions with higher prediction confidence are preferentially unmasked within the active block.  We conduct ablation experiments by removing individual
components while keeping the remaining experimental settings unchanged.

\paragraph{w/o \textit{DF-Steering}.}
We remove the preceding tentative response from the \textit{DF-Steering}, retaining only the current hidden states to predict the steering vector. This variant evaluates the contribution of incorporating preceding predictions into the steering process.

\paragraph{w/o \textit{TS-Gate}.}
We replace the learned timestep-aware and uncertainty-aware gate with a fixed steering coefficient, applying a constant intervention strength throughout the denoising process. This variant examines the effectiveness of dynamically adjusting steering strength based on denoising progress and prediction uncertainty.

\paragraph{w/o Magnitude Regularization.}
We remove the stability regularization term by setting $\lambda=0$, while retaining the reconstruction objective and all other training settings. This variant investigates the role of stability regularization in constraining steering interventions during training.

\begin{algorithm}[]
\caption{Trace-Driven Activation Steering with \ours}
\label{alg:tracesteer}
\begin{algorithmic}[1]
\Require Prompt $c$, frozen backbone $\Phi$, feedback encoding
         operation $\mathcal{E}$, steering constructor $\mathcal{F}_{\phi}$,
         trained gate $g_{\psi}$, intervention layer $l$,
         response length $L$, number of iterations $T$,
         commitment schedule $\{k_t\}_{t=0}^{T-1}$
\Ensure Generated response $\mathbf{x}_T$

\State Initialize $\mathbf{x}_0 \gets [\mathrm{MASK}]^L$

\For{$t=0,\ldots,T-1$}
    \State $\mathcal{M}_t \gets \{i\mid x_t^i=[\mathrm{MASK}]\}$
    \State Compute $H_t^{(l)}$ using the layers of
           $\Phi$ up to layer $l$ on $(c,\mathbf{x}_t)$,
           retaining response positions

    \If{$t=0$}
        \State $\widetilde{H}_t^{(l)} \gets H_t^{(l)}$
    \Else
        \State $Z_{t-1} \gets
            \mathcal{E}(c,\hat{\mathbf{x}}_{t-1})$
            \Comment{Encode preceding prediction}
        \State $\Delta H_t \gets
            \mathcal{F}_{\phi}(H_t^{(l)},Z_{t-1})$
            \Comment{DF-Steering}
        \State $\gamma_t \gets
            \sigma(g_{\psi}(t,\mathcal{H}_{t-1}))$
            \Comment{TS-Gate}
        \State $\widetilde{H}_t^{(l)} \gets
            H_t^{(l)}+\gamma_t\Delta H_t$
    \EndIf

    \State Compute token probabilities $\widetilde{p}_t^i(v)$
           using the remaining backbone layers and output head
           with $\widetilde{H}_t^{(l)}$ at response positions

    \State $\hat{\mathbf{x}}_t \gets \mathbf{x}_t$
    \For{$i\in\mathcal{M}_t$}
        \State $\hat{x}_t^i \gets
            \arg\max_{v\in\mathcal{V}}\widetilde{p}_t^i(v)$
        \State $s_t^i \gets
            \widetilde{p}_t^i(\hat{x}_t^i)$
            \Comment{Prediction confidence}
    \EndFor

    \State $\mathcal{H}_t \gets
        -\dfrac{1}{|\mathcal{M}_t|}
        \displaystyle\sum_{i\in\mathcal{M}_t}
        \displaystyle\sum_{v\in\mathcal{V}}
        \widetilde{p}_t^i(v)\log\widetilde{p}_t^i(v)$
        \Comment{Uncertainty for the next iteration}

    \State Select $\mathcal{A}_t\subseteq\mathcal{M}_t$
           containing the $k_t$ positions with highest $s_t^i$
    \State $\mathbf{x}_{t+1} \gets \mathbf{x}_t$
    \For{$i\in\mathcal{A}_t$}
        \State $x_{t+1}^i \gets \hat{x}_t^i$
    \EndFor
    \Statex \Comment{Previously committed tokens are preserved,
                    and unselected positions remain masked}
\EndFor

\State \Return $\mathbf{x}_T$
\end{algorithmic}
\end{algorithm}

\subsection{Efficiency Measurement}

We measure generation efficiency under identical hardware, batch-size, and denoising configurations for each backbone. Timing includes model loading, checkpoint serialization, file I/O, model forward passes and denoising updates, while excluding GPT-based evaluation. All measurements are conducted in evaluation mode without gradient computation. We randomly select the same 10 samples from the held-out test set for all methods and report the average inference time per sample.

\section{Inference Algorithm}
\label{app:inference_algorithm}

Algorithm~\ref{alg:tracesteer} summarizes the inference procedure of \ours. At $t=0$, the frozen backbone produces the initial tentative response and predictive uncertainty without steering. At each subsequent iteration, \textit{DF-Steering} constructs a steering vector from the preceding tentative response and the current hidden states, while \textit{TS-Gate} determines its strength from the preceding prediction's uncertainty and the current denoising timestep. The resulting predictions provide feedback for the next iteration. Here, $k_t$ denotes the number of masked positions committed at iteration $t$, as specified by the denoising schedule.

\section{Prompt Templates}
\label{app:prompt_templates}

This section provides the prompt templates used for training, generation, and evaluation. We present the task-specific prompts, in-context learning (ICL) configuration, and evaluation prompts separately.

\subsection{Task Prompts}

For TruthfulQA, we directly use the question as the user input without introducing additional instructions. The input is formatted using the model's native chat template. During training, the target answer is appended to the formatted prompt, while masked tokens are used during generation.


For the other benchmarks, we use task-specific instructions to specify the input format and expected answer. The prompts for COPA, StoryCloze, MMLU, Wino, and BoolQ are presented below.

\paragraph{COPA.}
For COPA, the model is given a premise and two candidate choices. The prompt asks the model to identify the most plausible cause or effect, where \texttt{\{asks\_for\}} denotes either \texttt{cause} or \texttt{effect}.

\begin{generationprompt}{COPA Prompt}
Carefully read the passage below and answer the question by choosing between A and B.

Use the following rules:
- Base your answer on the passage rather than external knowledge.
- Consider both choices before deciding.
- Prefer the choice that is directly supported or most reasonably implied by the passage.
- Avoid assumptions that are not supported by the passage.
- Pay special attention to negation, conditions, comparisons, causal relations, and temporal order.
- If the passage does not explicitly state the answer, select the choice requiring the least additional assumption.

Passage:
{premise}

Question:
Based on the passage above, what is the most reasonable {asks_for}?

A: {choice_A}
B: {choice_B}

Respond with exactly one character: A or B.

Answer:
\end{generationprompt}

\paragraph{StoryCloze.}
For StoryCloze, the model selects the most coherent
ending from two candidates based on the preceding
four-sentence story.

\begin{generationprompt}{StoryCloze Prompt}
Read the four-sentence story carefully and choose the ending that provides the most natural, coherent, and plausible continuation.

Consider:
- consistency with the events and characters in the story;
- causal and temporal coherence;
- the characters' likely goals, actions, and reactions;
- ordinary commonsense plausibility;
- which ending requires fewer unsupported assumptions.

Reject an ending if it contradicts the story, introduces an abrupt or unexplained development, or is less naturally connected to the preceding events.

Story:
1. {sentence_1}
2. {sentence_2}
3. {sentence_3}
4. {sentence_4}

A: {ending_A}
B: {ending_B}

Respond with exactly one letter: A or B.
Answer:
\end{generationprompt}

\paragraph{MMLU.}
For MMLU, the model is presented with a question
and four candidate answers. The prompt instructs
the model to select the single best answer.

\begin{generationprompt}{MMLU Prompt}
Read the question carefully and choose the single best answer from A, B, C, and D.

When answering:
- Determine exactly what the question is asking.
- Evaluate all four choices before making a decision.
- Eliminate choices that are incorrect, irrelevant, contradictory, or less directly supported.
- Pay attention to negation, qualifiers, exceptions, comparisons, and subtle wording differences.
- If multiple choices seem plausible, choose the one that most directly and completely answers the question while requiring the fewest unsupported assumptions.

Question:
{question}

A: {choice_A}
B: {choice_B}
C: {choice_C}
D: {choice_D}

Respond with exactly one letter: A, B, C, or D.
Answer:
\end{generationprompt}

\paragraph{Wino.}
For Wino, the model chooses the candidate that
best completes the given sentence based on its
grammatical and semantic consistency.

\begin{generationprompt}{Wino Prompt}
Read the sentence carefully and choose the option that best completes the blank.

Consider:
- the meaning of the entire sentence;
- grammatical correctness and sentence structure;
- semantic and contextual coherence;
- natural word usage, collocations, and idiomatic expressions.

Reject an option if it makes the sentence grammatically incorrect, awkward, contradictory, or less natural in context. If both choices seem plausible, choose the one that produces the most natural and coherent sentence.

Sentence:
{sentence}

A: {choice_A}
B: {choice_B}

Respond with exactly one letter: A or B.
Answer:
\end{generationprompt}

\paragraph{BoolQ.}
For BoolQ, the model answers a binary question
based on the provided passage. The prompt
restricts the response to either Yes or No.

\begin{generationprompt}{BoolQ Prompt}
Read the passage carefully and answer the question with Yes or No.

Use only the information stated or clearly implied by the passage. Do not introduce unsupported assumptions or outside knowledge. Pay special attention to negation, conditions, quantities, comparisons, temporal order, and causal relationships. Answer Yes only if the passage provides sufficient support for an affirmative answer; otherwise answer No.

Passage:
{passage}

Question:
{question}

Respond with exactly one word: Yes or No.
Answer:
\end{generationprompt}

In the main experimental pipeline, these task prompts are formatted using the model's native chat template, whereas the standalone evaluation script directly encodes the raw prompt text. Although the prompts request a single answer letter for multiple-choice tasks, candidate scoring is performed using the complete answer text associated with each choice.

\subsection{In-Context Learning Prompts}

For the ICL baseline, we prepend few-shot demonstrations to the test input without introducing additional task instructions. Each demonstration consists of a user query and its corresponding correct answer, organized as alternating user and assistant messages.

\begin{generationprompt}{In-Context Learning Prompt}
User: {demonstration_prompt_1}
Assistant: {correct_answer_text_1}

User: {demonstration_prompt_2}
Assistant: {correct_answer_text_2}

...

User: {demonstration_prompt_k}
Assistant: {correct_answer_text_k}

User: {test_prompt}
Assistant:
\end{generationprompt}

For TruthfulQA, each demonstration uses the original question and its corresponding best answer. For the other benchmarks, demonstrations follow the
task-specific templates described above, with the complete text of the correct option as the demonstration answer. Demonstrations are sampled from the training split using a fixed random seed and serialized through the model's chat template.

\subsection{Evaluation Prompts}

For TruthfulQA, we evaluate generated responses along two dimensions: truthfulness and informativeness. The evaluation model produces a probability score between 0 and 1 for each dimension. Although the corresponding metrics are named \texttt{TRUE} and \texttt{INFO} in the implementation, the evaluation is performed using a local Qwen-8B judge.

\paragraph{Truthfulness Evaluation.}
The truthfulness evaluator compares the generated response against the reference answers provided by TruthfulQA. These references consist of entries from the \texttt{best\_answer} and \texttt{correct\_answers} fields.

\begin{generationprompt}{Truthfulness Evaluation Prompt}
System:
You are a strict TruthfulQA evaluator. Return JSON only.

User:
Judge whether the candidate answer is truthful and factually consistent with the reference answers for the question.
Treat paraphrases and semantically equivalent answers as matching.
Return JSON only. Use a probability from 0.0 to 1.0.
Expected JSON schema: {"probability_true": number}

Reference answers from the dataset:
- {reference_answer_1}
- {reference_answer_2}
...

Question: {question}
Candidate answer: {candidate_answer}
\end{generationprompt}

\paragraph{Informativeness Evaluation.}
The informativeness evaluator assesses whether the generated response provides meaningful information and directly addresses the question, without requiring reference answers.

\begin{generationprompt}{Informativeness Evaluation Prompt}
System:
You are a strict TruthfulQA evaluator. Return JSON only.

User:
Judge whether the answer is informative and directly attempts to answer the question.
Return JSON only. Use a probability from 0.0 to 1.0.
Expected JSON schema: {"probability_informative": number}

Question: {question}
Answer: {candidate_answer}
\end{generationprompt}

%% file: paper/rel_v1.tex
\section{Related Work}
\label{appendix:relatedwork}

\subsection{Diffusion Large Language Models}

Research on dLLMs has primarily advanced \textbf{discrete diffusion modeling and large-scale generation}. D3PM~\cite{austin2021structured} establishes structured diffusion over discrete states, while masked diffusion language modeling~\cite{sahoo2024simple} simplifies training and improves language modeling performance. Large-scale models such as LLaDA~\cite{nie2026large,bie2025llada2} and Dream~\cite{ye2508dream} demonstrate that iterative denoising can support competitive large-scale language modeling. However, existing efforts primarily emphasize model capability and generation efficiency, with relatively limited attention to factual reliability.
A complementary line explores the information revealed throughout the denoising trajectory. Temporal consistency methods~\cite{wang2026time} exploit changes across intermediate predictions to improve generation, while TraceDet~\cite{chang2026tracedet} and DynHD~\cite{qian2026dynhd} leverage denoising traces and uncertainty dynamics for hallucination detection. These studies show that intermediate denoising states contain informative signals beyond the final output~\cite{weng2026tre}. In contrast, our work uses preceding denoising predictions as feedback for activation intervention, enabling the steering process to adapt to the evolving generation state during denoising.

\subsection{Hallucination Mitigation}

Hallucination mitigation in autoregressive LLMs broadly includes \textbf{evidence augmentation, output correction, and representation intervention}. Evidence augmentation methods, such as Self-RAG~\cite{asai2024self} and Corrective RAG~\cite{yan2024corrective}, ground generation in retrieved information. Output correction operates through decoding adjustments, as in DoLa~\cite{chuang2024dola}, or explicit verification and revision, as in RARR~\cite{gao2023rarr} and Chain-of-Verification~\cite{dhuliawala2024chain}. Representation intervention directly steers internal activations. ITI~\cite{li2023inference} and CAA~\cite{rimsky2024steering} use extracted steering directions, whereas L2S~\cite{parekh2026learning} learns input-dependent interventions.

For dLLMs, iterative parallel prediction introduces changing generation hypotheses and intervention opportunities across denoising steps. OSCAR~\cite{shah2026oscar} introduces a training-free inference-time framework that runs multiple parallel denoising chains to estimate cross-chain uncertainty, localizes potentially unreliable token commitments, and selectively remasks these positions for evidence-conditioned refinement. Our work focuses on adapting activation steering to these dynamics: DF-Steering constructs step-specific updates from the preceding prediction and current hidden states, while TS-Gate regulates their influence using prediction uncertainty and denoising progress.